\documentclass{article}
\usepackage[margin=1in]{geometry}
\usepackage{times}
\usepackage{hyperref}
\usepackage{url}
\usepackage{graphicx}
\usepackage{amsmath}
\usepackage{amsthm}
\usepackage{subcaption}
\usepackage{algorithm}
\usepackage{algorithmic}
\usepackage{natbib}

\newtheorem{theorem}{Theorem}
\newtheorem{definition}{Definition}
\DeclareMathOperator*{\argmax}{arg\,max}

\title{Online Structure Learning for Sum-Product Networks with Gaussian Leaves}

\author{
Wilson Hsu, Agastya Kalra \& Pascal Poupart \\
David R. Cheriton School of Computer Science\\
University of Waterloo\\
Waterloo, Ontario, Canada \\
\texttt{\{wwhsu,a6kalra,ppoupart\}@uwaterloo.ca} \\
}

\date{}

\begin{document}

\maketitle

\begin{abstract}
Sum-product networks have recently emerged as an attractive representation due to their dual view as a special type of deep neural network with clear semantics and a special type of probabilistic graphical model for which inference is always tractable. Those properties follow from some conditions (i.e., completeness and decomposability) that must be respected by the structure of the network.  As a result, it is not easy to specify a valid sum-product network by hand and therefore structure learning techniques are typically used in practice.  This paper describes the first {\em online} structure learning technique for continuous SPNs with Gaussian leaves. We also introduce an accompanying new parameter learning technique.
\end{abstract}

\section{Introduction}

Sum-product networks (SPNs) were first introduced by \citet{poon2011sum} as a new type of deep representation.  They distinguish themselves from other types of neural networks by several desirable properties: 
\begin{enumerate}
\item The quantities computed by each node can be clearly interpreted as (un-normalized) probabilities.
\item SPNs are equivalent to Bayesian and Markov networks~\citep{zhao2015spnbn} while ensuring that exact inference has linear complexity with respect to the size of the network.
\item They represent generative models that naturally handle arbitrary queries with missing data while changing which variables are treated as inputs and outputs.
\end{enumerate}
There is a catch: these nice properties arise only when the structure of the network satisfies certain conditions (i.e., decomposability and completeness)~\citep{poon2011sum}. Hence, it is not easy to specify sum-product networks by hand.  In particular, fully connected networks typically violate those conditions.  Similarly, most sparse structures that are handcrafted by practitioners to compute specific types of features or embeddings also violate those conditions. While this may seem like a major drawback, the benefit is that researchers have been forced to develop structure learning techniques to obtain valid SPNs that satisfy those conditions~\citep{dennis2012learning,gens2013learning,peharz2013greedy,lee2013online,rooshenas2014learning,adel2015learning,vergari2015simplifying,rahman2016merging,melibari2016dynamic}. At the moment, the search for good network structures in other types of neural networks is typically done by hand based on intuitions as well as trial and error.  However the expectation is that automated structure learning techniques will eventually dominate.  For this to happen, we need structure learning techniques that can scale easily to large amounts of data.  

To that effect, we propose the first {\em online} structure learning technique for SPNs with Gaussian leaves.  The approach starts with a network structure that assumes that all variables are independent. This network structure is then updated as a stream of data points is processed.  Whenever a statistically significant correlation is detected between some variables, a correlation is introduced in the network in the form of a multivariate Gaussian or a mixture distribution. This is done while ensuring that the resulting network structure is necessarily valid.  The approach is evaluated on several large benchmark datasets.

The paper is structured as follows.  Section~\ref{sec:background} provides some background about sum-product networks. 
Section~\ref{sec:algorithm} describes our {\em online} structure learning technique for SPNs with Gaussian leaves.  Section~\ref{sec:experiments} evaluates the performance of our structure learning technique on several large benchmark datasets.  Finally, Section~\ref{sec:conclusion} concludes the paper and discusses possible directions for future work.

\section{Background}
\label{sec:background}

Sum-product networks (SPNs) were first proposed by \citet{poon2011sum} as a new type of deep architecture consisting of a rooted acyclic directed graph with interior nodes that are sums and products while the leaves are tractable distributions, including Bernoulli distributions for discrete SPNs and Gaussian distributions for continuous SPNs.  The edges emanating from sum nodes are labeled with non-negative weights $w$. An SPN encodes a function $f(\bf{X}=\bf{x})$ that takes as input a variable assignment $\bf{X}=\bf{x}$ and produces an output at its root.  This function is defined recursively at each node $n$ as follows:
\begin{align}
f_n(\bf{X}=\bf{x}) = \left\{ \begin{array}{ll} 
\Pr(\bf{X}_n=\bf{x}_n) & \mbox{if $isLeaf(n)$} \\
\sum_{i} w_i f_{child_i(n)}(\bf{x}) & \mbox{if $isSum(n)$} \\
\prod_{i} f_{child_i(n)}(\bf{x}) & \mbox{if $isProduct(n)$}
\end{array} \right.
\end{align}
Here, $\bf{X}_n = \bf{x}_n$ denotes the variable assignment restricted to the variables contained in the leaf $n$. If none of the variables in leaf $n$ are instantiated by $\bf{X}=\bf{x}$ then $\Pr(\bf{X}_n=\bf{x}_n)=\Pr(\emptyset)=1$. Note also that if leaf $n$ contains continuous variables, then $\Pr(\bf{X}_n=\bf{x}_n)$ should be interpreted as $pdf(X_n=x_n)$.

An SPN is a neural network in the sense that each interior node can be interpreted as computing a linear combination of its children followed by a potentially non-linear activation function.  Without loss of generality, assume that the SPN is organized in alternating layers of sums and product nodes.\footnote{Consecutive sum nodes can always be merged into a single sum node. Similarly, consecutive product nodes can always be merged into a single product node.}  It is easy to see that sum-nodes compute a linear combination of their children.  Product nodes can be interpreted as the sum of its children in the log domain. Hence sum-product networks can be viewed as neural networks with logarithmic and exponential activation functions.  

An SPN can also be viewed as encoding a joint distribution over the random variables in its leaves when the network structure satisfies certain conditions.  These conditions are often defined in terms of the notion of {\em scope}.  

\begin{definition}[Scope] The $scope(n)$ of a node $n$ is the set of variables that are descendants of $n$.  
\end{definition}
A sufficient set of conditions to ensure a valid joint distribution includes:
\begin{definition}[Completeness~\citep{poon2011sum}]
An SPN is complete \textit{if} all children of the same sum node have the same scope.
\end{definition}
\begin{definition}[Decomposability~\citep{poon2011sum}]
An SPN is decomposable \textit{if} all children of the same product node have disjoint scopes.
\end{definition}
Here decomposability allows us to interpret product nodes as computing factored distributions with respect to disjoint sets of variables, which ensures that the product is a valid distribution over the union of the scopes of the children. Similarly, completeness allows us to interpret sum nodes as computing a mixture of the distributions encoded by the children since they all have the same scope.  Each child is a mixture component with mixture probability proportional to its weight.  Hence, in complete and decomposable SPNs, the sub-SPN rooted at each node can be interpreted as encoding an (un-normalized) joint distribution over its scope.  We can use the function $f$ to answer inference queries with respect to the joint distribution encoded by the entire SPN as follows:
\begin{itemize}
\item Marginal queries: $\Pr(\bf{X}=\bf{x}) = \frac{f_{root}(\bf{X}=\bf{x})}{f_{root}(\emptyset)}$ 
\item Conditional queries: $\Pr(\bf{X}=\bf{x}|\bf{Y}=\bf{y}) = \frac{f_{root}(\bf{X}=\bf{x},\bf{Y}=\bf{y})}{f_{root}(\bf{Y}=\bf{y})}$ 
\end{itemize}
Unlike most neural networks that can answer only queries with fixed inputs and outputs, SPNs can answer conditional inference queries with varying inputs and outputs simply by changing the set of variables that are queried (outputs) and conditioned on (inputs). Furthermore, SPNs can be used to generate data by sampling from the joint distributions they encode.  This is achieved by a top-down pass through the network.  Starting at the root, each child of a product node is followed, a single child of a sum node is sampled according to the unnormalized distribution encoded by the weights of the sum node and a variable assignment is sampled in each leaf that is reached. This is particularly useful in natural language generation tasks and image completion tasks~\citep{poon2011sum}.

Note also that inference queries can be answered exactly in linear time with respect to the size of the network since each query requires two evaluations of the network function $f$ and each evaluation is performed in a bottom-up pass through the network. This means that SPNs can also be viewed as a special type of tractable probabilistic graphical model, in contrast to Bayesian and Markov networks for which inference is \#P-hard~\citep{roth1996hardness}.  Any SPN can be converted into an equivalent bipartite Bayesian network without any exponential blow up, while Bayesian and Markov networks can be converted into equivalent SPNs at the risk of an exponential blow up~\citep{zhao2015spnbn}.

\subsection{Parameter Learning}

The weights of an SPN are its parameters.  They can be estimated by maximizing the likelihood of a dataset (generative training)~\citep{poon2011sum} or the conditional likelihood of some output features given some input features (discriminative training) by Stochastic Gradient Descent (SGD)~\citep{gens2012discriminative}.  Since SPNs are generative probabilistic models where the sum nodes can be interpreted as hidden variables that induce a mixture, the parameters can also be estimated by Expectation Maximization (EM)~\citep{poon2011sum,peharz2015foundations}. \cite{zhao2016unified} provides a unifying framework that explains how likelihood maximization in SPNs corresponds to a signomial optimization problem where SGD is a first order procedure, one can also consider a sequential monomial approximation and EM corresponds to a concave-convex procedure that converges faster than the other techniques.  Since SPNs are deep architectures, SGD and EM suffer from vanishing updates and therefore "hard" variants have been proposed to remedy to this problem~\citep{poon2011sum,gens2012discriminative}.  By replacing all sum nodes by max nodes in an SPN, we obtain a max-product network where the gradient is constant (hard SGD) and latent variables become deterministic (hard EM).  It is also possible to train SPNs in an online fashion based on streaming data~\citep{lee2013online,rashwan2016online,zhao2016collapsed,jaini2016}.  In particular, it was shown that online Bayesian moment matching~\citep{rashwan2016online,jaini2016} and online collapsed variational Bayes~\citep{zhao2016collapsed} perform much better than SGD and online EM.

\subsection{Structure Learning}

Since it is difficult to specify network structures for SPNs that satisfy the decomposability and completeness properties, several automated structure learning techniques have been proposed~\citep{dennis2012learning,gens2013learning,peharz2013greedy,lee2013online,rooshenas2014learning,adel2015learning,vergari2015simplifying,rahman2016merging,melibari2016dynamic}.  The first two structure learning techniques~\citep{dennis2012learning,gens2013learning} are top down approaches that alternate between instance clustering to construct sum nodes and variable partitioning to construct product nodes.  We can also combine instance clustering and variable partitioning in one step with a rank-one submatrix extraction by performing a singular value decomposition~\citep{adel2015learning}. Alternatively, we can learn the structure of SPNs in a bottom-up fashion by incrementally clustering correlated variables~\citep{peharz2013greedy}. These algorithms all learn SPNs with a tree structure and univariate leaves.  It is possible to learn SPNs with multivariate leaves by using a hybrid technique that learns an SPN in a top down fashion, but stops early and constructs multivariate leaves by fitting a tractable probabilistic graphical model over the variables in each leaf~\citep{rooshenas2014learning,vergari2015simplifying}. It is also possible to merge similar subtrees into directed acyclic graphs in a post-processing step to reduce the size of the resulting SPN~\citep{rahman2016merging}.  Furthermore, \cite{melibari2016dynamic} proposed dynamic SPNs for variable length data and described a search-and-score structure learning technique that does a local search over the space of network structures. 

So far, all these structure learning algorithms are batch techniques that assume that the full dataset is available and can be scanned multiple times. \cite{lee2013online} describes an online structure learning technique that gradually grows a network structure based on mini-batches.  The algorithm is a variant of LearnSPN~\citep{gens2013learning} where the clustering step is modified to use online clustering.  As a result, sum nodes can be extended with more children when the algorithm encounters a mini-batch that is better clustered with additional clusters. Product nodes are never modified after their creation. 

Since existing structure learning techniques have all been designed for discrete SPNs and have yet to be extended to continuous SPNs such as Gaussian SPNs, the state of the art for continuous (and large scale) datasets is to generate a random network structure that satisfies decomposability and completeness after which the weights are learned by a scalable online learning technique~\citep{jaini2016}.  We advance the state of the art by proposing a first online structure learning technique for Gaussian SPNs.

\section{Proposed Algorithm}
\label{sec:algorithm}

In this work, we assume that the leaf nodes all have Gaussian distributions. A
leaf node may have more than one variable in its scope, in which case it follows
a multivariate Gaussian distribution.

Suppose we want to model a probability distribution over a $d$-dimensional space.
The algorithm starts with a fully factorized joint probability distribution
over all variables, $p({\bf x}) = p(x_1, x_2, \ldots, x_d) = p_1(x_1)p_2(x_2) \cdots p_d(x_d)$.
This distribution is represented by a product node with $d$ children, the $i$th of
which is a univariate distribution over the variable $x_i$. Therefore, initially
we assume that the variables are independent, and the algorithm will update
this probability distribution as new data points are processed.

Given a mini-batch of data points, the algorithm
passes the points through the network from the root to the leaf nodes and updates each
node along the way. This update includes two parts:
\begin{itemize}
\item updating the parameters of the SPN, and
\item updating the structure of the network.
\end{itemize}

\subsection{Parameter update}

The parameters are updated by keeping track of running sufficient statistics. There are two
types of parameters in the model: weights on the branches under
a sum node, and parameters for the Gaussian distribution in a leaf node.

We propose a new online algorithm for parameter learning that is simple while ensuring that after each update, the likelihood of the last processed data point is increased (similar to stochastic gradient ascent). Algorithm~\ref{alg:param-update} describes the pseudocode of this procedure. Every node in the network has a count, $n_{c},$ initialized to 1. When a data point is received, the likelihood of this data point is computed at each node. Then the parameters of the network are updated in a recursive top-down fashion by starting at the root node.  When a sum node is traversed, its count is increased by 1 and the count of the child with the highest likelihood is increased by 1. This effectively increases the weight of the child with the highest likelihood while decreasing the weights of the remaining children. As a result, the overall likelihood at the sum node will increase. The weight $w_{s,c}$ of a branch between a sum node $s$ and one of its children $c$ can
then be estimated as
\begin{equation}
w_{s,c} = \frac{n_c}{n_s}
\end{equation}
where $n_s$ is the count of the sum node and $n_c$ is the count of the child node. We also recursively update the subtree of the child with the highest likelihood.  In the case of ties, we simply choose one of the children with highest likelihood at random to be updated.

\begin{algorithm}[!ht]
\caption{parameterUpdate(root(SPN),data)}
\label{alg:param-update}
\begin{algorithmic} 
\REQUIRE SPN and $m$ data points 
\ENSURE SPN with updated parameters
\STATE $n_{root} \leftarrow n_{root} + m$
\IF{$isProduct(root)$}
	\FOR{each $child$ of $root$}
		\STATE $parameterUpdate(child,data)$ 
    \ENDFOR
\ELSIF{$isSum(root)$}
	\FOR{each $child$ of $root$}
    	\STATE $subset \leftarrow \{x \in data \; | \; likelihood(child,x) \ge likelihood(child',x) \; \forall child' \; {\tt of} \; root\}$
		\STATE $parameterUpdate(child,subset)$
        \STATE $w_{root,child} \leftarrow \frac{n_{child}+1}{n_{root}+\#children}$
    \ENDFOR
\ELSIF{$isLeaf(root)$}
	\STATE update mean $\mu^{(root)}$ based on Eq.~\ref{eq:mean}
	\STATE update covariance matrix $\Sigma^{(root)}$ based on Eq.~\ref{eq:covariance}
\ENDIF
\end{algorithmic}
\end{algorithm}

Since there are no parameters associated with a product node, the only way to increase its likelihood is to increase the likelihood at each of its children.  We increment the count at each child of a product node and recursively update the subtrees rooted at each child. 

Since each leaf node represents a Gaussian distribution, it keeps track of the
empirical mean vector $\mu$ and empirical covariance matrix $\Sigma$ for the variables in its scope. When a leaf node with a current count of $n$ receives a batch of $m$ data points
$x^{(1)}, x^{(2)}, \ldots, x^{(m)}$, the empirical mean and empirical covariance are updated according to the equations:
\begin{equation}
\mu'_i = \frac{1}{n+m} \left(n \mu_i + \sum_{k=1}^m x^{(k)}_i\right)
\label{eq:mean}
\end{equation}
and
\begin{equation}
\Sigma'_{i,j} = \frac{1}{n+m}\left[n \Sigma_{i,j} + \sum_{k=1}^{m} \left(x_i^{(k)}-\mu_i\right)\left(x_j^{(k)}-\mu_j\right)\right] - (\mu'_i - \mu_i)(\mu'_j - \mu_j)
\label{eq:covariance}
\end{equation}
where $i$ and $j$ index the variables in the leaf node's scope, and $\mu'$ and $\Sigma'$ 
are the new mean and covariance after the update.

This parameter update technique is related to, but different from hard SGD and hard EM used in~\citep{poon2011sum,gens2012discriminative,lee2013online}.  Hard SGD and hard EM also keep track of a count for the child of each sum node and increment those counts each time a data point reaches this child.  However, to decide when a child is reached by a data point, they replace all descendant sum nodes by max nodes and evaluate the resulting max-product network. In contrast, we retain the descendant sum nodes and evaluate the original sum-product network as it is.  This evaluates more faithfully the probability that a data point is generated by a child.

Alg.~\ref{alg:param-update} does a single pass through the data.  The complexity of updating the parameters after each data point is linear in the size of the network (i.e., \# of edges) since it takes one bottom up pass to compute the likelihood of the data point at each node and one top-down pass to update the sufficient statistics and the weights. The update of the sufficient statistics can be seen as locally maximizing the likelihood of the data.  The empirical mean and covariance of the Gaussian leaves locally increase the likelihood of the data that reach that leaf.  Similarly, the count ratios used to set the weights under a sum node locally increase the likelihood of the data that reach each child. We prove this result below.

\begin{theorem}
Let $\theta_s$ be the set of parameters of an SPN $s$, and let $f_s(\cdot|\theta_s)$ be the probability density function of the SPN.
Given an observation $x$, suppose the parameters are updated to $\theta'_s$ based on the running average update procedure, then
we have $f_s(x|\theta'_s) \geq f_s(x|\theta_s)$.
\end{theorem}

\emph{Proof.} We will prove the theorem by induction. First suppose the SPN is just one leaf node. In this case, the parameters are the empirical mean and covariance, which is the maximum likelihood estimator for Gaussian distribution. Suppose $\theta$ consists of the parameters learned using $n$ data points $x^{(1)}, \ldots, x^{(n)}$, and $\theta'$ consists of the parameters learned using the same $n$ data points and an additional observation $x$. Then we have
\begin{equation}
f_s(x|\theta'_s) \prod_{i=1}^nf(x^{(i)}|\theta'_s) \geq f_s(x|\theta_s) \prod_{i=1}^nf_s(x^{(i)}|\theta_s) \geq f_s(x|\theta_s) \prod_{i=1}^nf_s(x^{(i)}|\theta'_s)
\end{equation}
Thus we get $f_s(x|\theta'_s) \geq f_s(x|\theta_s)$.

Now suppose we have an SPN $s$ where each child SPN $t$ satisfies the property $f_t(x|\theta'_t) \geq f_t(x|\theta_t)$. If the root of $s$ is a product node, then $f_s(x|\theta'_s) = \prod_t f_t(x|\theta'_t) \geq \prod_t f_t(x|\theta_t) = f_s(x|\theta_s)$.

Now suppose the root of $s$ is a sum node. Let $n_t$ be the count of child $t$, and let $u = \argmax_t f_t(x|\theta_t)$. Then we have
\begin{align*}
f_s(x|\theta'_s) &= \frac{1}{n+1} \left(f_u(x|\theta'_u) + \sum_t n_t f_t(x|\theta'_t)\right) \\
&\geq \frac{1}{n+1} \left(f_u(x|\theta_u) + \sum_t n_t f_t(x|\theta_t)\right) ~~~\textrm{by inductive hypothesis} \\
&\geq \frac{1}{n+1} \left(\sum_t \frac{n_t}{n} f_t(x|\theta_t) + \sum_t n_t f_t(x|\theta_t)\right) \\
&= \frac{1}{n} \sum_t n_t f_t(x|\theta_t) \\
&= f_s(x|\theta_s) 
\qed
\end{align*}

\subsection{Structure update}
\label{sec:structupdate}

The simple online parameter learning described above can be easily extended to enable online structure learning.  Algorithm~\ref{alg:oSLRAU} describes the pseudocode of the resulting procedure called oSLRAU (online Structure Learning with Running Average Update). Similar to leaf nodes, each product node also keeps track of the empirical mean vector and
empirical covariance matrix of the variables in its scope. These are updated in the same
way as the leaf nodes.

Initially, when a product node is created, all variables in the scope are assumed
independent (see Algorithm~\ref{alg:createFactoredModel}). As new data points arrive at a product node, the covariance matrix
is updated, and if the absolute value of the Pearson correlation coefficient
between two variables are above a certain threshold, the algorithm updates the
structure so that the two variables become correlated in the model.

\begin{figure}
\centering
\includegraphics[width=0.95\textwidth]{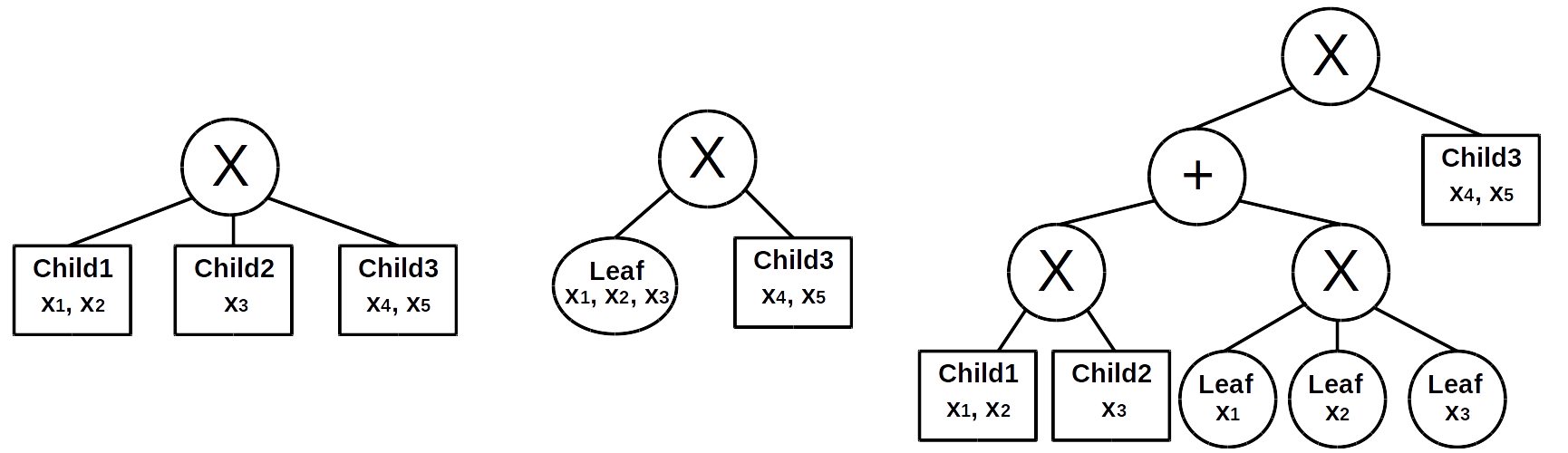}
\caption{Depiction of how correlations between variables are introduced in the model.
Left: original product node with three children. Middle: combine Child1 and Child2
into a multivariate leaf node (Alg.~\ref{alg:createMultiVarGaussian}). Right: create a mixture to model the correlation (Alg.~\ref{alg:createMixture}).}
\label{fig:merge}
\end{figure}

We correlate two variables in the model by combining the child nodes
whose scopes contain the two variables. The algorithm employs two approaches to
combine the two child nodes:
\begin{itemize}
\item create a multivariate leaf node (Algorithm~\ref{alg:createMultiVarGaussian}), or
\item create a mixture of two components over the variables (Algorithm~\ref{alg:createMixture}).
\end{itemize}
These two processes are depicted in Figure~\ref{fig:merge}. On the left, a product
node with scope $x_1, \ldots, x_5$ originally has three children. The product node
keeps track of the empirical mean and empirical covariance for these five variables. Suppose it
receives a mini-batch of data and updates the statistics.  As a result of this update,
$x_1$ and $x_3$ now have a correlation above the threshold.

Figure~\ref{fig:merge} illustrates the two approaches to model this correlation.  In the middle of Figure~\ref{fig:merge}, the algorithm combines the two child nodes that have $x_1$
and $x_3$ in their scope, and turns them into a multivariate leaf node. Since the product
node already keeps track of the mean and covariance of these variables, we can
simply use those statistics as the parameters for the new leaf node.

Another way to correlate $x_1$ and $x_3$ is to create a mixture, as shown in the right part of Figure~\ref{fig:merge}. The mixture has two components. The first component contains
the original children of the product node that contain $x_1$ and $x_3$. The second component is a new product
node, which is again initialized to have a fully factorized distribution over its
scope (Alg.~\ref{alg:createFactoredModel}). The mini-batch of data points are then passed down the new mixture to update
its parameters.

Note that although the children are drawn like leaf nodes in the diagrams, they can
in fact be entire subtrees. Since the process does not involve the parameters in
a child, it works the same way if some of the children are trees instead of single
nodes.

The technique chosen to induce a correlation depends on the number of variables in the scope.
The algorithm creates a multivariate leaf node when the combined scope of the
two child nodes has a number of variables that does not exceed some threshold and if the total number of variables
in the problem is greater than this threshold, otherwise it creates a mixture. Since the number of parameters in multivariate Gaussian leaves grows at a quadratic rate with respect to the number of variables, it is not advised to consider multivariate leaves with too many variables.  In contrast, the mixture construction increases the number of parameters at a linear rate, which is less prone to overfitting when many variables are correlated.  

To simplify the structure, if a product node ends up with only one child, it
is removed from the network, and its only child is joined with its parent.
Similarly, if a sum node ends up being a child of another sum node, then the
child sum node can be removed, and all its children are promoted one layer up.

Note that the this structure learning technique does a single pass through the data and therefore is entirely online. The time and space complexity of updating the structure after each data point is linear in the size of the network (i.e., \# of edges) and quadratic in the number of features (since product nodes store a covariance matrix that is quadratic in the size of their scope).  The algorithm also ensures that the decomposability and completeness properties are preserved after each update.

Our algorithm (oSLRAU) is related to, but different from the online structure learning technique proposed by \cite{lee2013online}.  Lee et al.'s technique was applied to discrete datasets while oSLRAU learns SPNs with Gaussian leaves based on real-valued data.  Furthermore, Lee et al.'s technique incrementally constructs a network in a top down fashion by adding children to sum nodes by online clustering.  Once a product node is constructed, it is never modified.  In contrast, oSLRAU incrementally constructs a network in a bottom up fashion by detecting correlations and modifying product nodes to represent these correlations.  Finally, Lee et al.'s technique updates the parameters by hard EM (which implicitly works with a max-product network) while oSLRAU updates the parameters by Alg.~\ref{alg:param-update} (which retains the original sum-product network) as explained in the previous section.    

\begin{algorithm}[!ht]
\caption{$oSLRAU(root(SPN),data)$}
\label{alg:oSLRAU}
\begin{algorithmic} 
\REQUIRE SPN and $m$ data points 
\ENSURE SPN with updated parameters
\STATE $n_{root} \leftarrow n_{root} + m$
\IF{$isProduct(root)$}
	\STATE update covariance matrix $\Sigma^{(root)}$ based on Eq.~\ref{eq:covariance}
    \STATE $highestCorrelation \leftarrow 0$
    \FOR{each $c,c'\in children(root)$ where $c\neq c'$}
    	\STATE $correlation_{c,c'} \leftarrow \max_{i\in scope(c),j\in scope(c')} \frac{|\Sigma_{ij}^{(root)}|}{\sqrt{\Sigma_{ii}^{(root)}\Sigma_{jj}^{(root)}}}$
 		\IF{$correlation_{c,c'} > highestCorrelation$}
        	\STATE $highestCorrelation \leftarrow correlation_{c,c'}$
            \STATE $child_1 \leftarrow c$
            \STATE $child_2 \leftarrow c'$
        \ENDIF
    \ENDFOR
    \IF{$highest \ge threshold$}
    	\IF{$|scope(child_1)\cup scope(child_2)| \ge nVars$}
        	\STATE $createMixture(root,child_1,child_2)$
        \ELSE
        	\STATE $createMultivariateGaussian(root,child_1,child_2)$
        \ENDIF
    \ENDIF
	\FOR{each $child$ of $root$}
		\STATE $oSLRAU(child,data)$ 
    \ENDFOR
\ELSIF{$isSum(root)$}
	\FOR{each $child$ of $root$}
    	\STATE $subset \leftarrow \{x \in data \; | \; likelihood(child,x) \ge likelihood(child',x) \; \forall child' \; {\tt of} \; root\}$
		\STATE $oSLRAU(child,subset)$
        \STATE $w_{root,child} \leftarrow \frac{n_{child}+1}{n_{root}+\#children}$
    \ENDFOR
\ELSIF{$isLeaf(root)$}
	\STATE update mean $\mu^{(root)}$ based on Eq.~\ref{eq:mean}
	\STATE update covariance matrix $\Sigma^{(root)}$ based on Eq.~\ref{eq:covariance}
\ENDIF
\end{algorithmic}
\end{algorithm}

\begin{algorithm}[!ht]
\caption{$createMixture(root,child_1,child_2)$}
\label{alg:createMixture}
\begin{algorithmic} 
\REQUIRE SPN and two children to be merged 
\ENSURE new mixture model
\STATE remove $child_1$ and $child_2$ from $root$
\STATE $component_1 \leftarrow$ create product node
\STATE add $child_1$ and $child_2$ as children of $component_1$
\STATE $n_{component_1} \leftarrow n_{root}$
\STATE $jointScope \leftarrow scope(child_1) \cup scope(child_2)$
\STATE $\Sigma^{(component_1)} \leftarrow \Sigma^{(root)}_{jointScope,jointScope}$
\STATE $component_2 \leftarrow createFactoredModel(jointScope)$
\STATE $n_{component_2} \leftarrow 0$
\STATE $mixture \leftarrow$ create sum node
\STATE add $component_1$ and $component_2$ as children of $mixture$
\STATE $n_{mixture} \leftarrow n_{root}$
\STATE $w_{mixture,component_1} \leftarrow \frac{n_{component_1}+1}{n_{mixture}+2}$
\STATE $w_{mixture,component_2} \leftarrow \frac{n_{component_2}+1}{n_{mixture}+2}$
\STATE add $mixture$ as child of $root$
\STATE return $root$
\end{algorithmic}
\end{algorithm}

\begin{algorithm}[!ht]
\caption{$createMultiVarGaussian(root,child_1,child_2)$}
\label{alg:createMultiVarGaussian}
\begin{algorithmic} 
\REQUIRE SPN, two children to be merged and data 
\ENSURE new multivariate Gaussian
\STATE create $multiVarGaussian$
\STATE $jointScope \leftarrow \{ scope(child_1) \cup scope(child_2)\}$
\STATE $\mu^{(multiVarGaussian)} \leftarrow \mu^{(root)}_{jointScope}$
\STATE $\Sigma^{(multiVarGaussian)} \leftarrow \Sigma^{(root)}_{jointScope,jointScope}$
\STATE $n_{multiVarGaussian} \leftarrow n_{root}$
\STATE return $multiVarGaussian$
\end{algorithmic}
\end{algorithm}

\begin{algorithm}[!ht]
\caption{$createFactoredModel(scope)$}
\label{alg:createFactoredModel}
\begin{algorithmic} 
\REQUIRE scope (set of variables) 
\ENSURE fully factored SPN
\STATE $factoredModel \leftarrow$ create product node
\FOR{each $i \in scope$}
	\STATE add $N_i(\mu{=}0,\sigma{=}\Sigma^{(root)}_{i,i})$ as child of $factoredModel$
\ENDFOR
\STATE $\Sigma^{(factoredModel)} \leftarrow {\bf 0}$
\STATE $n_{factoredModel} \leftarrow 0$
\STATE return $factoredModel$
\end{algorithmic}
\end{algorithm}

\section{Experiments}
\label{sec:experiments}

This section discusses our experiments to evaluate the performance of our structure learning technique.\footnote{The source code for our algorithm is available at {\tt github.com/whsu/spn}.}

\subsection{Toy dataset}

As a proof of concept, we first test the algorithm on a toy synthetic dataset.
We generate data from the 3-dimensional distribution 
\begin{align*}
p(x_1, x_2, x_3) = [ &0.25 N(x_1|1,1)N(x_2|2,2) + 0.25 N(x_1|11,1)N(x_2|12,2) \\
&+ 0.25 N(x_1|21,1) N(x_2|22,2) + 0.25 N(x_1|31,1) N(x_2|32,2) ] N(x_3|3,3),
\end{align*}
where $N(\cdot|\mu,\sigma^2)$ is the normal distribution with mean $\mu$ and variance $\sigma^2$.

Therefore, the first two dimensions $x_1$ and $x_2$ are generated from a Gaussian mixture with four components, and $x_3$ is independent from the other two variables.

Starting from a fully factorized distribution, we would expect $x_3$ to remain factorized
after learning from data. Furthermore, the algorithm should generate new components along the first two dimensions as more data
points are received since $x_1$ and $x_2$ are correlated.

\begin{figure}
\centering
\includegraphics[width=0.95\textwidth]{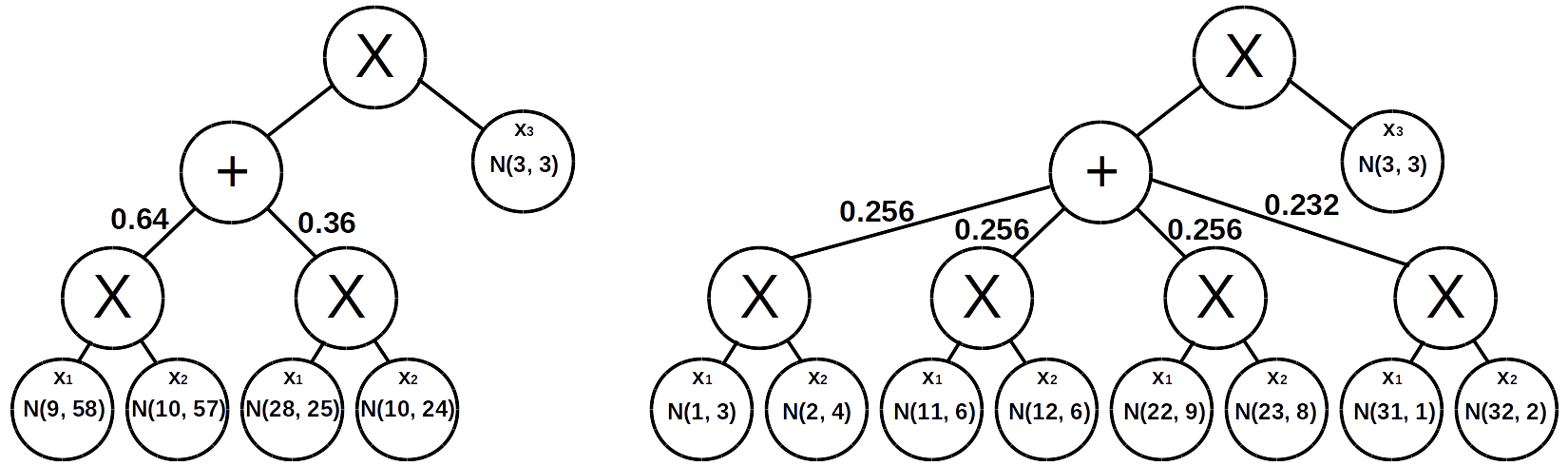}
\caption{Learning the structure from the toy dataset using univariate leaf nodes.
Left: after 200 data points. Right: after 500 data points.}
\label{fig:toy}
\end{figure}

This is indeed what happens. Figure~\ref{fig:toy} shows the structure learned after 
200 and 500 data points. The variable $x_3$ remains factorized regardless of the number
of data points seen, whereas more components are created for $x_1$ and $x_2$ as more
data points are processed.

\begin{figure}
\centering
\includegraphics[width=0.95\textwidth]{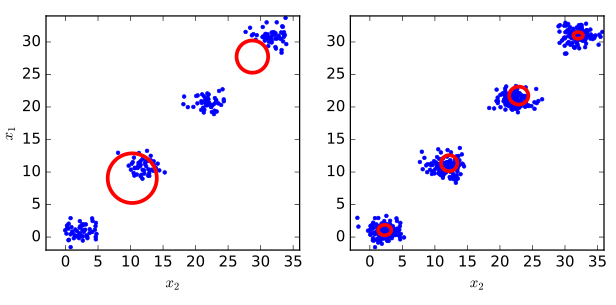}
\caption{Blue dots are the data points from the toy dataset, and the red ellipses show the diagonal Gaussian components learned. Left: after 200 data points. Right: after 500 data points.}
\label{fig:toydata}
\end{figure}

Figure~\ref{fig:toydata} shows the data points along the first two dimensions and
the Gaussian components learned. We can see that the algorithm generates new
components to model the correlation between $x_1$ and $x_2$ as it processes more data.

\subsection{Comparison to other Algorithms}

In a second experiment, we compare our algorithm to several alternatives on the same datasets used by~\cite{jaini2016}. We use 0.1 as the correlation threshold in all experiments, and we use mini-batch sizes of 1 for the three datasets with fewest instances (Quake,
Banknote, Abalone), 8 for the two slightly larger ones (Kinematics, CA), and 256
for the two datasets with most instances (Flow Size, Sensorless).

The experimental results for our algorithm called {\em online structure learning with
running average update} (oSLRAU) are listed in Table~\ref{tab:loglhd} along with
results reproduced from~\cite{jaini2016}. The table reports the average test
log likelihoods with standard error on 10-fold cross validation.  oSLRAU achieved better log likelihoods than online Bayesian moment matching (oBMM)~\citep{jaini2016} and online expectation maximization (oEM)~\citep{cappe2009line} with network structures generated at random or corresponding to Gaussian mixture models (GMMs). This highlights the main advantage of oSLRAU: learning a structure that models the data. Stacked Restricted Boltzmann Machines (SRBMs)~\citep{salakhutdinov2009deep} and Generative Moment Matching Networks (GenMMNs)~\citep{li2015} are other types of deep generative models. Since it is not possible to compute the likelihood of data points with GenMMNs, the model is augmented with Parzen windows.  More specifically, 10,000 samples are generated using the resulting GenMMNs and a Gaussian kernel is estimated for each sample by adjusting its parameters to maximize the likelihood of a validation set.  However, as pointed out by~\cite{theis2015} this
method only provides an approximate estimate of the log-likelihood and therefore the log-likelihood
reported for GenMMNs in Table~\ref{tab:loglhd} may not be directly comparable to the log-likelihood of other models.

The network structures for GenMMNs and SRBMs are fully connected while ensuring that the number of parameters is comparable to those of the SPNs.  oSLRAU outperforms these models on 5 datasets while SRBMs and GenMMNs each outperform oSLRAU on one dataset. Although SRBMs and GenMMNs are more expressive than SPNs since they allow other types of nodes beyond sums and products, training GenMMNs and SRBMs is notoriously difficult.  In contrast, oSLRAU provides a simple and effective way of optimizing the structure and parameters of SPNs that captures well the correlations between variables and therefore yields good results. 

\begin{table}
\small
\centering
\caption{Average log-likelihood scores with standard error on small real-world data sets. The best results are
highlighted in bold. (random) indicates a random network structure and (GMM) indicates a fixed network structure corresponding to a Gaussian mixture model.}
\label{tab:loglhd}
\begin{tabular}{|c|c|c|c|c|c|c|c|}
\hline
Dataset   & Flow Size & Quake & Banknote & Abalone & Kinematics & CA & Sensorless \\
\# of vars &  3   &   4   &     4    &    8    &      8     & 22 &    48      \\
\hline
\hline
oSLRAU & \textbf{14.78} & \textbf{-1.86} & -2.04 & \textbf{-1.12} & -11.15 & \textbf{17.10} & \textbf{54.82} \\
& $\pm$ 0.97 & $\pm$ 0.20 & $\pm$ 0.15 & $\pm$ 0.21 & $\pm$ 0.03 & $\pm$ 1.36 & $\pm$ 1.67 \\
\hline
\hline
oBMM & - & - & - & ~-1.82 & -11.19 & ~-2.47 & ~1.58 \\
(random) & & & & $\pm$ 0.19 & $\pm$ 0.03 & $\pm$ 0.56 & $\pm$ 1.28 \\
\hline
oEM & - & - & - & -11.36 & -11.35 & -31.34 & -3.40 \\
(random) & & & & $\pm$ 0.19 & $\pm$ 0.03 & $\pm$ 1.07 & $\pm$ 6.06 \\
\hline
\hline
oBMM & 4.80 & -3.84 & -4.81 & -1.21 & -11.24 & -1.78 & - \\
(GMM) & $\pm$ 0.67 & $\pm$ 0.16 & $\pm$ 0.13 & $\pm$ 0.36 & $\pm$ 0.04 & $\pm$ 0.59 &  \\
\hline
oEM & -0.49 & -5.50 & -4.81 & -3.53 & -11.35 & -21.39 & - \\
(GMM) & $\pm$ 3.29 & $\pm$ 0.41 & $\pm$ 0.13 & $\pm$ 1.68 & $\pm$ 0.03 & $\pm$ 1.58 &  \\
\hline
\hline
SRBM & -0.79 & -2.38 & -2.76 & -2.28 & \textbf{-5.55} & -4.95 & -26.91 \\
& $\pm$ 0.004 & $\pm$ 0.01 & $\pm$ 0.001 & $\pm$ 0.001 & $\pm$ 0.02 & $\pm$ 0.003 & $\pm$ 0.03 \\
\hline
GenMMN & 0.40 & -3.83 & \textbf{-1.70} & -3.29 & -11.36 & -5.41 & -29.41 \\
& $\pm$ 0.007 & $\pm$ 0.21 & $\pm$ 0.03 & $\pm$ 0.10 & $\pm$ 0.02 & $\pm$ 0.14 & $\pm$ 1.16 \\
\hline
\end{tabular}
\end{table}

\subsection{Large Datasets}

We also tested oSLRAU on larger datasets to evaluate its scaling properties.  Table~\ref{table:info-large-datasets} shows the number of attributes and data points in each dataset. Table~\ref{table:likelihood-large-datasets} compares the average log-likelihood of oSLRAU to that of randomly generated networks (which are the state of the art for obtain a valid continuous SPNs) for those large datasets.  For a fair comparison we generated random networks that are at least as large as the networks obtained by oSLRAU. oSLRAU achieves higher log-likelihood than random networks since it effectively discovers empirical correlations and generates a structure that captures those correlations. 

\begin{table}[]
\centering
\caption{Information for each large dataset}
\label{table:info-large-datasets}
\begin{tabular}{|l|r|r|r|r|}
\hline
\textbf{Dataset} & \textbf{Datapoints} & \textbf{Variables} \\ \hline 
Voxforge         & 3,603,643           & 39                \\ \hline 
Power            & 2,049,280           & 4                  \\ \hline
Network          & 434,873             & 3                  \\ \hline
GasSen           & 8,386,765           & 16                \\ \hline 
MSD              & 515,344             & 90                \\ \hline 
GasSenH          & 928,991             & 10                \\ \hline 
\end{tabular}
\end{table}

\begin{table}[]
\centering
\caption{Average log-likelihood scores with standard error on large real-world data sets. The best results among the online techniques (random, oSLRAU and RealNVP online) are
highlighted in bold.  Results for RealNVP offline are also included for comparison purposes.}
\label{table:likelihood-large-datasets}
\begin{tabular}{|l|c|c|c|c|}
\hline
\textbf{Datasets} & \textbf{Random}  & \textbf{oSLRAU} & \textbf{RealNVP Online} & \textbf{RealNVP Offline}\\ \hline
Voxforge          & -33.9 $\pm$ 0.3   & {\bf -29.6} $\pm$ 0.0  & -169.0 $\pm$ 0.6 & -168.2 $\pm$ 0.8 \\ \hline
Power             & -2.83 $\pm$ 0.13  & {\bf -2.46} $\pm$ 0.11  & -18.70 $\pm$ 0.19 & -17.85 $\pm$ 0.22 \\  \hline
Network           & -5.34 $\pm$ 0.03  & {\bf -4.27} $\pm$ 0.04 & -10.80 $\pm$ 0.02 & -7.89 $\pm$ 0.05\\ \hline
GasSen            & -114 $\pm$ 2  & {\bf -102} $\pm$ 4 & -748 $\pm$ 99 & -443 $\pm$ 64 \\ \hline
MSD               & -538.8 $\pm$ 0.7 & -531.4 $\pm$ 0.3 & {\bf -362.4} $\pm$ 0.4 & -257.1 $\pm$ 2.03 \\ \hline
GasSenH           & -21.5 $\pm$ 1.3   & {\bf -15.6} $\pm$ 1.2 & -44.5 $\pm$ 0.1 & 44.2 $\pm$ 0.1 \\ \hline
\end{tabular}
\end{table}

We also compare oSLRAU to a publicly available implementation of RealNVP\footnote{https://github.com/taesung89/real-nvp}. Since the benchmarks include a variety of problems from different domains and it is not clear what network architecture would work best, we used a default 2-hidden-layer fully connected network. The two layers have the same size. For a fair comparison, we used a number of nodes per layer that yields approximately the same number of parameters as the sum product networks.  Training was done by stochastic gradient descent in TensorFlow with a step size of 0.01 and mini-batch sizes that vary from 100 to 1500 depending on the size of the dataset. We report the results 
for online learning (single iteration) and offline learning (validation loss stops decreasing). In this experiment, the correlation threshold was kept constant at 0.1. To determine the maximum number of variables in multivariate leaves, we followed the following rule: at most one variable per leaf if the problem has 3 features or less and then increase the maximum number of variables per leaf up to 4 depending on the number of features. Further analysis on the effects of varying the maximum number of variables per leaf are available below. We do this to balance the size and the expressiveness of the resulting SPN. oSLRAU outperformed RealNVP on 5 of the 6 datasets.  This can be explained by the fact that oSLRAU learns a structure that is suitable for each problem while RealNVP does not learn any structure. Note that it should be possible for RealNVP to obtain better results by using a better architecture than a default 2-hidden-layer network, however in the absence of domain knowledge this is difficult.  Furthermore, in online learning with streaming data, it is not possible to do an offline search over some hyperparameters such as the number of layers and nodes in order to fine tune the architecture.  Hence, the results presented in Table~\ref{table:likelihood-large-datasets} highlight the importance of an online structure learning technique such as oSLRAU to obtain a suitable network structure with streaming data in the absence of domain knowledge.

Table~\ref{table:early-stopping} reports the training time (seconds) and the size (\# of nodes) of the resulting SPNs for each dataset when running oSLRAU and a variant that stops structure learning early.  The experiments were carried out on an Amazon c4.xlarge machine with 4 vCPUs (high frequency Intel Xeon E5-2666 v3 Haswell processors) and 7.5 Gb of RAM.  The times are relatively short since oSLRAU is an online algorithm and therefore does a single pass through the data. Since it gradually constructs the structure of the SPN as it processes the data, we can also stop the updates to the structure early (while still updating the parameters).   This helps to mitigate overfitting while producing much smaller SPNs and reducing the running time. In the columns labeled "early stop" we report the results achieved when structure learning is stopped after processing one ninth of the data. The resulting SPNs are significantly smaller, while achieving a log-likelihood that is close to that of oSLRAU without early stopping.

\begin{table}[]
\centering
\caption{Large datasets: comparison of oSLRAU with and without early stopping (i.e., no structure learning after one ninth of the data is processed, but still updating the parameters).}
\label{table:early-stopping}
\begin{tabular}{|l|c|c|r|r|r|r|r|}
\hline
 & \multicolumn{2}{|c|}{\textbf{log-likelihood}} & \multicolumn{2}{|c|}{\textbf{time (sec)}} & \multicolumn{2}{|c|}{\textbf{SPN size (\# nodes)}} \\ \hline
\textbf{Dataset} & \textbf{oSLRAU} & \textbf{early stop} & \textbf{oSLRAU} & \textbf{early stop} & \textbf{oSLRAU} & \textbf{early stop} \\ \hline
Power   & -2.46 $\pm$ 0.11            & \textbf{-0.24} $\pm$ 0.20 & 183 & 70 & 23360 & 1154 \\ \hline
Network & \textbf{-4.27} $\pm$ 0.02 & -4.30 $\pm$ 0.02          & 14 & 4 & 7214 & 249\\ \hline
GasSen  & \textbf{-102} $\pm$ 4 & -111 $\pm$ 3          & 351 & 188 & 5057 & 564 \\ \hline
MSD     & {\bf -531.4} $\pm$ 0.3         & -534.9 $\pm$ 0.3          & 44 & 26 & 672 & 238  \\ \hline
GasSenH & {\bf -15.6} $\pm$ 1.2          & -18.6 $\pm$ 1.0           & 12 & 9 & 920 & 131 \\ \hline
\end{tabular}
\end{table}

\begin{table}[]
\centering
\caption{Log likelihoods with standard error as we vary the threshold for the maximum \# of variables in a multivariate Gaussian leaf. No results are reported (dashes) when the maximum \# of variables is greater than the total number of variables.}
\label{table:leaf-node-likelihood}
\begin{tabular}{|l|c|c|c|c|c|}
\hline
                & \multicolumn{5}{c|}{\textbf{Maximum \# of Variables per Leaf Node}}                                                            \\ \hline
\textbf{Dataset} & 1                 & 2                 & 3                & 4                 & 5                    \\ \hline
 Power            & {\bf -1.71} $\pm$ 0.18   & -3.02 $\pm$ 0.24   & -3.74 $\pm$ 0.28  & -4.52 $\pm$ 0.1   & ------               \\ \hline
Network          & {\bf -4.27} $\pm$ 0.09   & -4.53 $\pm$ 0.09   & -4.75 $\pm$ 0.02  & ------            & ------               \\ \hline
GasSen           & -105 $\pm$ 2.5 & -103 $\pm$ 2.8 & {\bf -102} $\pm$ 4.1 & -104 $\pm$ 3.8 & -103 $\pm$ 3.8 \\ \hline
MSD              & -532 $\pm$ 0.32 & {\bf -531} $\pm$ 0.32 & {\bf -531} $\pm$ 0.28 & {\bf -531} $\pm$ 0.31 & -532 $\pm$ 0.34     \\ \hline
GasSenH          & -17.2 $\pm$ 1.04  & -16.8 $\pm$ 1.23  & \bf -15.6 $\pm$ 1.13 & -15.9 $\pm$ 1.3  & \b-16.1 $\pm$ 1.47 \\ \hline
\end{tabular}
\end{table}

\begin{table}[]
\centering
\caption{Average times (seconds) as we vary the threshold for the maximum \# of variables in a multivariate Gaussian leaf. No results are reported (dashes) when the maximum \# of variables is greater than the total number of variables.}
\label{table:leaf-node-time}
\begin{tabular}{|l|l|l|l|l|l|}
\hline
                & \multicolumn{5}{c|}{\textbf{Maximum \# of Variables per Leaf Node}}    \\ \hline
\textbf{Dataset} & 1      & 2      & 3      & 4      & 5       \\ \hline
Power            & 133 & 41.5 & 13.8 & 9.9    & ------  \\ \hline
Network          & 14.1 & 4.01  & 1.92  & ------ & ------  \\ \hline
GasSen           & 783.78 & 450.34 & 350.52 & 148.89 & 145.759 \\ \hline
MSD              & 80.47  & 64.44  & 44.9   & 43.65  & 41.44   \\ \hline
GasSenH          & 16.59 & 13.35 & 11.76 & 11.04 & 10.16   \\ \hline
\end{tabular}
\end{table}

\begin{table}[]
\centering
\caption{Average SPN sizes (\# of nodes) as we vary the threshold for the maximum \# of variables in a multivariate Gaussian leaf. No results are reported (dashes) when the maximum \# of variables is greater than the total number of variables.}
\label{table:leaf-node-size}
\begin{tabular}{|l|l|l|l|l|l|}
\hline
                & \multicolumn{5}{c|}{\textbf{Maximum \# of Variables per Leaf Node}} \\ \hline
\textbf{Dataset} & 1      & 2     & 3     & 4      & 5      \\ \hline
Power            & 14269   & 2813  & 427  & 8      & ------ \\ \hline
Network          & 7214   & 1033  & 7     & ------ & ------ \\ \hline
GasSen           & 13874  & 6879  & 5057  & 772    & 738    \\ \hline
MSD              & 6547   & 3114  & 802   & 672    & 582    \\ \hline
GasSenH          & 1901   & 1203  & 920  & 798   & 664 \\ \hline
\end{tabular}
\end{table}

The size of the resulting SPNs and their log-likelihood also depend on the correlation threshold used to determine when the structure should be updated to account for a detected correlation, and the maximum size of a leaf node used to determine when to branch off into a new subtree.

To understand the impact that the maximum number of variables per leaf node has on the resulting SPN, we performed experiments where the minibatch size and correlation threshold were held constant for a given dataset while the maximum number of variables per leaf node varies. We report the log likelihood with standard error after ten-fold cross validation, as well as  average size and average time in Tables~\ref{table:leaf-node-likelihood},~\ref{table:leaf-node-time} and~\ref{table:leaf-node-size}. As expected, the number of nodes in an SPN decreases as the leaf node cap increases, since there will be less branching. What's interesting is that depending on the type of correlations in the datasets, different sizes perform better or worse. For example in Power, we notice that univariate leaf nodes are the best, but in GasSenH, slightly larger leaf nodes tend to do well. We show that too many variables in a leaf node leads to worse performance and underfitting, and in some cases too few variables per leaf node leads to overfitting. These results show that in general, the largest decrease in size and time while maintaining good performance occurs with a maximum of 3 variables per leaf node. Therefore in practice, 3 variables per leaf node works well, except when there are only a few variables in the dataset, then 1 is a good choice. 

Tables~\ref{table:correlation-likelihood}, ~\ref{table:correlation-time} and~\ref{table:correlation-size} show respectively how the log-likelihood, time and size changes as we vary the correlation threshold from 0.05 to 0.7.  A very small correlation threshold tends to detect spurious correlations and lead to overfitting while a large correlation threshold tends to miss some correlations and lead to underfitting.  The results in Table~\ref{table:correlation-likelihood} generally support this tendency subject to noise due to sample effects.  Since the highest log-likelihood was achieved in three of the datasets with a correlation threshold of 0.1, this explains why we used 0.1 as the threshold in the previous experiments.  Tables~\ref{table:correlation-time} and~\ref{table:correlation-size} also show that the average time and size of the resulting SPNs generally decrease (subject to noise) as the correlation threshold increases since fewer correlations tend to be detected. 

\begin{table}[]
\centering
\caption{Log Likelihoods for different correlation thresholds.}
\label{table:correlation-likelihood}
\begin{tabular}{|l|c|c|c|c|c|c|}
\hline
                   & \multicolumn{6}{c|}{\textbf{Correlation Threshold}}                                                                   \\ \hline
\textbf{Dataset} & 0.05              & 0.1               & 0.2               & 0.3               & 0.5               & 0.7               \\ \hline \hline
Power              & -2.37 $\pm$ 0.13   & -2.46 $\pm$ 0.11   & {\bf -2.20} $\pm$ 0.18   & -3.02 $\pm$ 0.24   & -4.65 $\pm$ 0.11   & -4.68 $\pm$ 0.09   \\ \hline
Network            & {\bf -3.98} $\pm$ 0.09    & -4.27 $\pm$ 0.02   & -4.75 $\pm$ 0.02   & -4.75 $\pm$ 0.02   & -4.75 $\pm$ 0.02   & -4.75 $\pm$ 0.02   \\ \hline
GasSen             & -104 $\pm$ 5  & {\bf -102} $\pm$ 4  & {\bf -102} $\pm$ 3 & {\bf -102} $\pm$ 3  &     -103 $\pm$ 3 &       -110 $\pm$ 3 \\ \hline
MSD                & {\bf -531.4} $\pm$ 0.3 & {\bf -531.4} $\pm$ 0.3 & {\bf -531.4} $\pm$ 0.3 & {\bf -531.4} $\pm$ 0.3 & -532.0 $\pm$ 0.3 & -536.2 $\pm$ 0.1 \\ \hline
GasSenH            & {\bf -15.6} $\pm$ 1.2  & {\bf -15.6} $\pm$ 1.2  & -15.8 $\pm$ 1.1  & -16.2 $\pm$ 1.4  & -16.1 $\pm$ 1.4  & -17.2 $\pm$ 1.4  \\ \hline
\end{tabular}
\end{table}

\begin{table}[]
\centering
\caption{Average times (seconds) as we vary the correlation threshold.}
\label{table:correlation-time}
\begin{tabular}{|l|r|r|r|r|r|r|}
\hline
                 & \multicolumn{6}{c|}{\textbf{Correlation Threshold}} \\ \hline
\textbf{Dataset} & 0.05    & 0.1     & 0.2   & 0.3    & 0.5   & 0.7    \\ \hline \hline
Power            & 197   & 183   & 130 & 39  & 10   & 9    \\ \hline
Network          & 20    & 14   & 1.9  & 1.9   & 1.9  & 1.9   \\ \hline
GasSen           & 370  & 351  & 349 & 366 & 423 & 142 \\ \hline
MSD              & 44.3   & 43.7   & 44.3 & 44.0 & 43.0 & 30.3  \\ \hline
GasSenH          & 11.8  & 11.7 & 11.9   & 13.0 & 12.0  & 15.1 \\ \hline
\end{tabular}
\end{table}

\begin{table}[]
\centering
\caption{Average SPN sizes (\# of nodes) as the correlation threshold changes.}
\label{table:correlation-size}
\begin{tabular}{|l|r|r|r|r|r|r|}
\hline
                 & \multicolumn{6}{c|}{\textbf{Correlation Threshold}} \\ \hline
\textbf{Dataset} & 0.05     & 0.1      & 0.2    & 0.3   & 0.5   & 0.7  \\ \hline \hline
Power            & 24914  & 23360  & 16006  & 2813  & 11    & 11   \\ \hline
Network          & 11233    & 7214     & 9      & 9     & 9     & 9    \\ \hline
GasSen           & 5315     & 5057     & 5041   & 5035  & 4581  & 490  \\ \hline
MSD              & 672      & 672      & 674    & 674   & 660   & 448  \\ \hline
GasSenH          & 920    & 920  & 887     & 877   & 1275  & 796 \\ \hline
\end{tabular}
\end{table}

\section{Conclusion and Future work}
\label{sec:conclusion}

This paper describes a first online structure learning technique for Gaussian SPNs that does a single pass through the data.  This allowed us to learn the structure of Gaussian SPNs in domains for which the state of the art was previously to generate a random network structure. This algorithm can also scale to large datasets efficiently.   

In the future, this work could be extended in several directions.  We are investigating the combination of our structure learning technique with other
parameter learning methods. Currently, we are simply learning the parameters by
keeping running statistics for the weights, mean vectors, and covariance matrices.  It might be possible to improve the performance by using more sophisticated
parameter learning algorithms.
We would also like to extend the structure learning algorithm to discrete variables. Finally, we would like to look into ways to automatically control the complexity
of the networks. For example, it would be useful to add a regularization mechanism
to avoid possible overfitting.

\bibliography{main}

\begin{thebibliography}{22}
\providecommand{\natexlab}[1]{#1}
\providecommand{\url}[1]{\texttt{#1}}
\expandafter\ifx\csname urlstyle\endcsname\relax
  \providecommand{\doi}[1]{doi: #1}\else
  \providecommand{\doi}{doi: \begingroup \urlstyle{rm}\Url}\fi

\bibitem[Adel et~al.(2015)Adel, Balduzzi, and Ghodsi]{adel2015learning}
Tameem Adel, David Balduzzi, and Ali Ghodsi.
\newblock Learning the structure of sum-product networks via an svd-based
  algorithm.
\newblock In \emph{UAI}, 2015.

\bibitem[Capp{\'e} and Moulines(2009)]{cappe2009line}
Olivier Capp{\'e} and Eric Moulines.
\newblock On-line expectation--maximization algorithm for latent data models.
\newblock \emph{Journal of the Royal Statistical Society: Series B (Statistical
  Methodology)}, 71\penalty0 (3):\penalty0 593--613, 2009.

\bibitem[Dennis and Ventura(2012)]{dennis2012learning}
Aaron Dennis and Dan Ventura.
\newblock Learning the architecture of sum-product networks using clustering on
  variables.
\newblock In \emph{NIPS}, 2012.

\bibitem[Gens and Domingos(2012)]{gens2012discriminative}
Robert Gens and Pedro Domingos.
\newblock Discriminative learning of sum-product networks.
\newblock In \emph{NIPS}, pages 3248--3256, 2012.

\bibitem[Gens and Domingos(2013)]{gens2013learning}
Robert Gens and Pedro Domingos.
\newblock Learning the structure of sum-product networks.
\newblock In \emph{ICML}, pages 873--880, 2013.

\bibitem[Jaini et~al.(2016)Jaini, Rashwan, Zhao, Liu, Banijamali, Chen, and
  Poupart]{jaini2016}
Priyank Jaini, Abdullah Rashwan, Han Zhao, Yue Liu, Ershad Banijamali, Zhitang
  Chen, and Pascal Poupart.
\newblock Online algorithms for sum-product networks with continuous variables.
\newblock In \emph{International Conference on Probabilistic Graphical Models
  (PGM)}, 2016.

\bibitem[Lee et~al.(2013)Lee, Heo, and Zhang]{lee2013online}
Sang-Woo Lee, Min-Oh Heo, and Byoung-Tak Zhang.
\newblock Online incremental structure learning of sum--product networks.
\newblock In \emph{International Conference on Neural Information Processing
  (ICONIP)}, pages 220--227. Springer, 2013.

\bibitem[Li et~al.(2015)Li, Swersky, and Zemel]{li2015}
Yujia Li, Kevin Swersky, and Rich Zemel.
\newblock Generative moment matching networks.
\newblock In \emph{ICML}, pages 1718--1727, 2015.

\bibitem[Melibari et~al.(2016)Melibari, Poupart, Doshi, and
  Trimponias]{melibari2016dynamic}
Mazen Melibari, Pascal Poupart, Prashant Doshi, and George Trimponias.
\newblock Dynamic sum-product networks for tractable inference on sequence
  data.
\newblock In \emph{JMLR Conference and Workshop Proceedings - International
  Conference on Probabilistic Graphical Models (PGM)}, 2016.

\bibitem[Peharz(2015)]{peharz2015foundations}
Robert Peharz.
\newblock \emph{Foundations of Sum-Product Networks for Probabilistic
  Modeling}.
\newblock PhD thesis, Medical University of Graz, 2015.

\bibitem[Peharz et~al.(2013)Peharz, Geiger, and Pernkopf]{peharz2013greedy}
Robert Peharz, Bernhard~C Geiger, and Franz Pernkopf.
\newblock Greedy part-wise learning of sum-product networks.
\newblock In \emph{Machine Learning and Knowledge Discovery in Databases},
  pages 612--627. Springer, 2013.

\bibitem[Poon and Domingos(2011)]{poon2011sum}
Hoifung Poon and Pedro Domingos.
\newblock Sum-product networks: A new deep architecture.
\newblock In \emph{UAI}, pages 2551--2558, 2011.

\bibitem[Rahman and Gogate(2016)]{rahman2016merging}
Tahrima Rahman and Vibhav Gogate.
\newblock Merging strategies for sum-product networks: From trees to graphs.
\newblock In \emph{Proceedings of the Thirty-Second Conference on Uncertainty
  in Artificial Intelligence, UAI}, 2016.

\bibitem[Rashwan et~al.(2016)Rashwan, Zhao, and Poupart]{rashwan2016online}
Abdullah Rashwan, Han Zhao, and Pascal Poupart.
\newblock Online and {D}istributed {B}ayesian {M}oment {M}atching for
  {S}um-{P}roduct {N}etworks.
\newblock In \emph{AISTATS}, 2016.

\bibitem[Rooshenas and Lowd(2014)]{rooshenas2014learning}
Amirmohammad Rooshenas and Daniel Lowd.
\newblock Learning sum-product networks with direct and indirect variable
  interactions.
\newblock In \emph{ICML}, pages 710--718, 2014.

\bibitem[Roth(1996)]{roth1996hardness}
Dan Roth.
\newblock On the hardness of approximate reasoning.
\newblock \emph{Artificial Intelligence}, 82\penalty0 (1):\penalty0 273--302,
  1996.

\bibitem[Salakhutdinov and Hinton(2009)]{salakhutdinov2009deep}
Ruslan Salakhutdinov and Geoffrey~E Hinton.
\newblock Deep boltzmann machines.
\newblock In \emph{AISTATS}, pages 448--455, 2009.

\bibitem[Theis et~al.(2015)Theis, Oord, and Bethge]{theis2015}
Lucas Theis, A{\"a}ron Oord, and Matthias Bethge.
\newblock A note on the evaluation of generative models.
\newblock \emph{arXiv:1511.01844}, 2015.

\bibitem[Vergari et~al.(2015)Vergari, Di~Mauro, and
  Esposito]{vergari2015simplifying}
Antonio Vergari, Nicola Di~Mauro, and Floriana Esposito.
\newblock Simplifying, regularizing and strengthening sum-product network
  structure learning.
\newblock In \emph{ECML-PKDD}, pages 343--358. 2015.

\bibitem[Zhao and Poupart(2016)]{zhao2016unified}
Han Zhao and Pascal Poupart.
\newblock A unified approach for learning the parameters of sum-product
  networks.
\newblock \emph{arXiv:1601.00318}, 2016.

\bibitem[Zhao et~al.(2015)Zhao, Melibari, and Poupart]{zhao2015spnbn}
Han Zhao, Mazen Melibari, and Pascal Poupart.
\newblock On the relationship between sum-product networks and {Bayesian}
  networks.
\newblock In \emph{ICML}, 2015.

\bibitem[Zhao et~al.(2016)Zhao, Adel, Gordon, and Amos]{zhao2016collapsed}
Han Zhao, Tameem Adel, Geoff Gordon, and Brandon Amos.
\newblock Collapsed variational inference for sum-product networks.
\newblock In \emph{ICML}, 2016.

\end{thebibliography}
\bibliographystyle{plainnat}

\end{document}